\documentclass[10pt,letterpaper]{ieeetran}

\usepackage{amsmath,amsfonts,amssymb}
\usepackage{graphicx}
\usepackage{caption}
\usepackage{bm}
\usepackage{url}

\usepackage{color}
\def\zhao{\textcolor{black}}
\usepackage{titlesec}

\titlespacing*{\subsection}{0pt}{1.ex}{0ex}

\title{IBCapsNet: Information Bottleneck Capsule Network for Noise-Robust Representation Learning}

\author{Canqun Xiang$^{1}$, Chen Yang$^{2}$ and Jiaoyan Zhao$^{2}$$^{*}$

\thanks{Canqun Xiang is an independent researcher (e-mail: xiangcanqun2018@email.szu.edu.cn).}
\thanks{Chen Yang and Jiaoyan Zhao are with Institute of Applied Artificial Intelligence of the Guangdong-Hong Kong-Macao Greater Bay Area, Shenzhen Polytechnic University, Shenzhen, China (e-mail: yangchen@szpu.edu.cn; zhaojy@szpu.edu.cn).}
\thanks{$^{*}$Corresponding author}
}

\begin{document}

\maketitle

\begin{abstract}
\zhao{Capsule networks (CapsNets) are superior at modeling hierarchical spatial relationships} but suffer from two critical limitations: high computational cost due to iterative dynamic routing and poor robustness under input corruptions. To address these issues, we propose IBCapsNet, a novel capsule architecture grounded in the Information Bottleneck (IB) principle. Instead of iterative routing, \zhao{IBCapsNet employs a one-pass variational aggregation mechanism, where primary capsules are first compressed into a global context representation and then processed by class-specific variational autoencoders (VAEs) to infer latent capsules regularized by the KL divergence}. This design enables efficient inference while inherently filtering out noise. Experiments on MNIST, Fashion-MNIST, SVHN and CIFAR-10 show that IBCapsNet matches CapsNet in clean-data accuracy (achieving \zhao{99.41\% on MNIST and 92.01\%} on SVHN), yet significantly outperforms it under four types of synthetic noise—demonstrating average improvements of +17.10\% and +14.54\% for clamped additive and multiplicative noise, respectively. Moreover, IBCapsNet achieves 2.54$\times$ faster training and 3.64$\times$ higher inference throughput compared to CapsNet, while reducing model parameters by 4.66\%. Our work bridges information-theoretic representation learning with capsule networks, offering a principled path toward robust, efficient, and interpretable deep models. code is available at \url{https://github.com/cxiang26/IBCapsnet}
\end{abstract}


\section{Introduction}
\label{sec:intro}

\begin{figure}[pt]
    \centering
    \includegraphics[width=\linewidth]{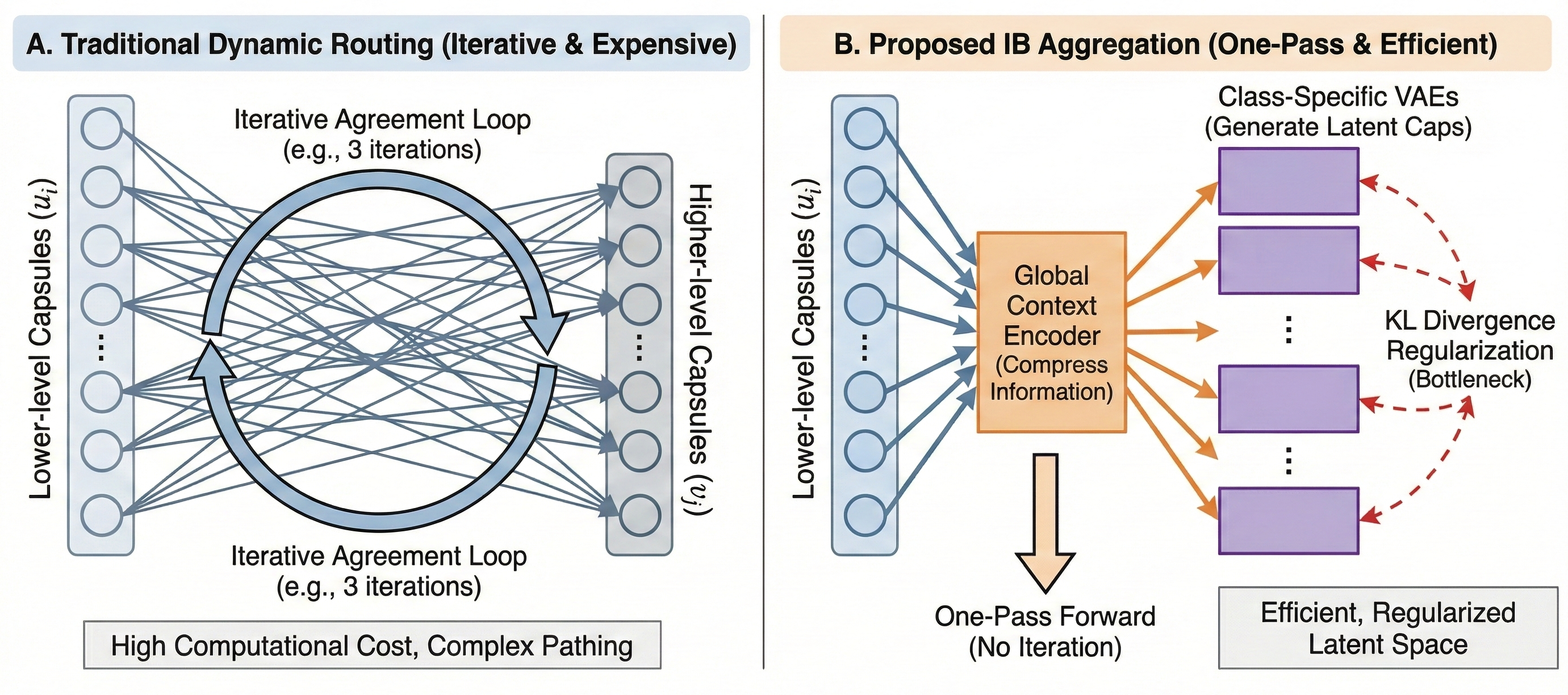}
    \caption{This figure contrasts the iterative, computationally expensive dynamic routing machanism (A) with the proposed efficient, one-pass Information Bottleneck aggregation machanism (B).}
    \label{fig:caps_vs_ibcapsv1}
\end{figure}
\IEEEPARstart{C}{apsule} networks (CapsNets) [1] have emerged as a promising alternative to conventional convolutional neural networks (CNNs) by explicitly modeling hierarchical pose relationships through vectorized capsule representations. A key component enabling this capability is the dynamic routing mechanism, which iteratively refines agreement-based coupling coefficients between lower- and higher-level capsules to ensure that only consistent features are routed upward.

However, \zhao{this iterative procedure incurs substantial computational overhead and is highly sensitive to input perturbations}. Dynamic routing relies on iterative agreement between the predicted parent capsules and the actual lower-level capsules, where the coupling coefficients $c_{ij}$ are updated based on similarities of the dot-product. When input corruptions (e.g. noise, or blur) distort the primary capsule activations, even small perturbations can break the delicate consensus required for stable routing, causing the algorithm to converge to suboptimal coupling patterns or fail to reach consensus altogether. This leads to error propagation through the hierarchy, ultimately degrading classification performance. Although several variants have attempted to alleviate these issues—e.g., by directing expectation maximization [2] or sparse attention [3]—they still fundamentally rely on local consistency assumptions that remain vulnerable to global distortions and do not explicitly address the information-theoretic question of \textit{what information should be retained} versus discarded.

In this work, we argue that robust capsule aggregation should be guided by an information-theoretic principle: retain only task-relevant information while discarding redundant or noisy components. The Information Bottleneck (IB) principle [4] provides a principled framework, positing that optimal representations should minimize mutual information with input $I(X; Z)$ while preserving task-relevant information $I(Z; Y)$. This naturally \zhao{exhibits noise-suppression capability}: by compressing the input representation through a bottleneck, the model is forced to discard irrelevant details (including noise) while retaining only discriminative features. Inspired by this principle, we propose IBCapsNet, a novel capsule architecture that replaces dynamic routing with a one-pass variational aggregation mechanism grounded in class-conditional generative modeling. Specifically, our model first encodes primary capsules into a compact global context representation, which is then fed into a set of class-specific variational autoencoders (VAEs) to infer latent capsules that balance reconstruction fidelity and distributional regularization via a KL divergence bottleneck, as shown in \textbf{Fig. ~\ref{fig:caps_vs_ibcapsv1}}. This design enables end-to-end training without any iterative routing, resulting in faster inference and representations that are inherently robust to input corruptions.

We conducted comprehensive experiments on MNIST, Fashion-MNIST, SVHN, and CIFAR-10 under four types of synthetic noise. The results show that IBCapsNet matches the accuracy of CapsNet on clean data, while significantly outperforming it and LeNet under all corruption conditions. Moreover, IBCapsNet produces remarkably stable reconstructions even under strong noise, whereas CapsNet's outputs vary erratically. In summary, our contributions are threefold: (i) We introduce the first capsule network grounded in the Information Bottleneck principle, replacing iterative routing with principled variational aggregation that explicitly models information compression through KL-divergence regularization; (ii) We demonstrate significant robustness gains across multiple datasets and noise types (e.g. +40.99\% improvement on MNIST under clamped additive noise), without sacrificing clean-data accuracy; (iii) We provide empirical and qualitative evidence through reconstruction stability and ablation studies that our design yields more reliable and interpretable capsule representations, with substantial computational efficiency gains (2.54$\times$ faster training, 3.64$\times$ higher inference throughput).

\section{Related Work}
\label{sec:related}

\textbf{Robust Capsule Networks.} Since the introduction of CapsNets \cite{Sabour2018capsnet}, numerous efforts have attempted to improve routing mechanisms, including EM routing \cite{hinton2018matrix}, attention-based routing \cite{zhang2018attention}, consensus mechanisms \cite{hahn2019capsule}, sparsity constraints \cite{geng2024orthcaps}, adversarial training \cite{shah2024adversarial}, multi-scale architectures \cite{xiang2018mscapsnet}, and matrix capsule methods \cite{xiang2020matrix}. However, these methods focus on \textit{how} to route information rather than \textit{what} information should be retained, and none explicitly model information compression or provide a principled mechanism for filtering noise.
\textbf{Information Bottleneck in Deep Learning.} The Information Bottleneck principle \cite{tishby1999information} provides a theoretical framework for learning compressed, task-relevant representations. Variational Information Bottleneck (VIB) \cite{alemi2017deep} enables tractable optimization of IB through variational inference, approximating mutual information terms with KL divergence. VIB has been successfully applied to classification, domain generalization \cite{hafezkolahi2019information}, and adversarial defense \cite{huang2021robust}, demonstrating that explicit information compression can improve robustness. Recent works extend IB principles to graph neural networks \cite{wu2020graph} and transformers \cite{jiang2020information}. However, these applications focus on unstructured or sequence-based representations. None explore IB within structured object representations where parts and wholes are explicitly modeled through hierarchical capsule hierarchies. To our knowledge, IBCapsNet is the first to combine VIB with CapsNets, addressing this gap by modeling class capsules as stochastic latent variables inferred from a compressed global context, enabling both structured representation and noise-robust compression.

\section{Method}
\label{sec:method}


\subsection{Preliminaries and Motivation}
Capsule networks \cite{Sabour2018capsnet} represent entities as activity vectors whose length signifies the probability of presence and orientation encodes the pose. In the standard formulation, lower-level capsules $\{\mathbf{u}_i\}_{i=1}^N$ are routed to higher-level capsules via dynamic routing—an iterative process that updates coupling coefficients $c_{ij}$ based on agreement between predicted and actual parent capsules. While effective on clean data, this mechanism is inherently fragile: small input corruptions distort low-level capsules, breaking the consensus required for stable routing and leading to error propagation. Moreover, the iterative nature incurs significant computational overhead.

To overcome these limitations, we advocate replacing agreement-based routing with an information-theoretic aggregation principle. The Information Bottleneck (IB) framework \cite{alemi2017deep} posits that optimal representations $Z$ should minimize mutual information $I(X; Z)$ with the input while preserving task-relevant information $I(Z; Y)$:
\begin{equation}
\min_{p(z|x)} I(X; Z) - \beta I(Z; Y)
\end{equation}
\zhao{where $\beta$ controls the trade-off between compression and task performance}. Applied to capsule hierarchies, this suggests that high-level capsules should be \textit{inferred} from a compressed, noise-robust summary rather than assembled through fragile local agreements. By enforcing compression through a bottleneck, the model is forced to discard noise and irrelevant details while retaining only discriminative features, enabling \textbf{one-pass, top-down aggregation} guided by class-specific generative priors.

\subsection{IBCapsNet Architecture}
IBCapsNet replaces iterative dynamic routing with a one-pass, information bottleneck–guided aggregation process. 
The network first encodes the input into primary capsules, then compresses them into a global context vector. Instead of routing via local agreement, it infers a set of class-specific latent capsules through parallel variational autoencoders (VAEs), each regularized by a KL-divergence bottleneck. Classification is performed using the norms of these latent capsules under the margin loss [1], while reconstruction leverages the winning capsule for input recovery. IBCapsNet consists of four key components:

\begin{enumerate}
    \item \textbf{Primary Capsule Layer}: An initial convolutional stack processes the input image $\mathbf{x}$ to produce $N$ primary capsules $\{\mathbf{u}_i \in \mathbb{R}^d\}_{i=1}^N$, identical to \cite{Sabour2018capsnet}.
    
    \item \textbf{Global Context Encoder}: All primary capsules are aggregated into a compact global context vector $\mathbf{h} \in \mathbb{R}^m$: 
    \begin{equation}
    \mathbf{h} = \text{MLP}\left(\left[\frac{1}{d}\sum_{j=1}^d u_{i,j}\right]_{i=1}^N\right)
    \end{equation}
    \zhao{where $u_{i,j}$ denotes the $j$-th component of capsule $\mathbf{u}_i \in \mathbb{R}^d$}. This step compresses each capsule by averaging its $d$ components, then applies a two-layer MLP to produce the global context vector, discarding spatial redundancy and enforcing a global bottleneck.
    
    \item \textbf{Class-Specific Variational Autoencoders (VAEs)}: For each class $c \in \{1, \dots, C\}$, a dedicated VAE infers a latent capsule $\mathbf{z}_c \in \mathbb{R}^k$ conditioned on $\mathbf{h}$. The encoder defines an approximate posterior:
    \begin{equation}
    \begin{aligned}
        q_{\phi_c}(\mathbf{z}_c \mid \mathbf{h}) 
        &= \mathcal{N}\bigl(\boldsymbol{\mu}_c, \operatorname{diag}(\boldsymbol{\sigma}_c^2)\bigr), \\
        &\text{where } 
        \bigl[\boldsymbol{\mu}_c, \log \boldsymbol{\sigma}_c^2\bigr] 
        = \operatorname{MLP}_c(\mathbf{h})
    \end{aligned}
    \end{equation}
    and the latent capsule is sampled via reparameterization:
    \begin{equation}
    \mathbf{z}_c = \boldsymbol{\mu}_c + \boldsymbol{\sigma}_c \odot \boldsymbol{\varepsilon}, \quad \boldsymbol{\varepsilon} \sim \mathcal{N}(0, \mathbf{I})
    \end{equation}
    This process is \textbf{non-iterative} and executed in parallel across all classes.
    
    \item \textbf{Classification and Reconstruction Heads}:
\begin{itemize}
    \item \textit{Classification}: The activity of class $c$ is represented by the norm of its latent capsule, $a_c = \|\mathbf{z}_c\|$. We adopt the margin loss from~\cite{Sabour2018capsnet}:
    \begin{equation}
    \begin{split}
        \mathcal{L}_{\text{cls}} = 
        &\sum_{c=1}^{C} T_c \bigl[\max(0, m^{+} - a_c)\bigr]^2 \\
        &+ \lambda (1 - T_c) \bigl[\max(0, a_c - m^{-})\bigr]^2
    \end{split}
    \end{equation}
    where $T_c = 1$ if class $c$ is the ground-truth label and $0$ otherwise, with $m^{+} = 0.9$, $m^{-} = 0.1$, and $\lambda = 0.5$.

    \item \textit{Reconstruction}: Given the predicted class $\hat{y} = \arg\max_{c} a_c$, a shared decoder reconstructs the input: $\hat{\mathbf{x}} = \mathrm{Decoder}(\mathbf{z}_{\hat{y}})$. The reconstruction loss is $\mathcal{L}_{\text{recon}} = \|\mathbf{x} - \hat{\mathbf{x}}\|_2^2$. The reconstruction module acts as a denoising signal that encourages the model to retain semantically meaningful features while discarding noise, particularly effective when combined with the KL bottleneck.
\end{itemize}
\end{enumerate}

\subsection{Training Objective}
IBCapsNet is trained end-to-end with the following composite loss:
\begin{equation}
\mathcal{L} = \mathcal{L}_{\text{cls}} + \lambda \mathcal{L}_{\text{recon}} + \beta \sum_{c=1}^C D_{\text{KL}}\left( q_{\phi_c}(\mathbf{z}_c | \mathbf{h}) \,\|\, p(\mathbf{z}_c) \right)
\end{equation}
where $\mathcal{L}_{\text{cls}}$ is the margin loss \cite{Sabour2018capsnet}, $\mathcal{L}_{\text{recon}} = \|\mathbf{x} - \hat{\mathbf{x}}\|_2^2$, and $p(\mathbf{z}_c) = \mathcal{N}(0, \mathbf{I})$ is a standard Gaussian prior. The KL term acts as an \textbf{information bottleneck}, regularizing each latent capsule to retain only class-discriminative structure while suppressing noise. Minimizing $D_{\text{KL}}(q_{\phi_c}(\mathbf{z}_c | \mathbf{h}) \| p(\mathbf{z}_c))$ encourages the posterior to be close to the prior, effectively compressing the information content and forcing the model to discard redundant and noisy information while retaining only task-relevant features. The trade-off parameter $\beta$ controls the strength of this compression.

\section{Experiments}
\label{sec:experiments}
We conduct comprehensive experiments to evaluate IBCapsNet across four datasets (MNIST, Fashion-MNIST, CIFAR-10, and SVHN) under four types of synthetic noise. We compare against standard CapsNet~\cite{Sabour2018capsnet} and LeNet-5~\cite{lenet} as baselines. All models use identical preprocessing, training protocols, and hyperparameters for fair comparison.

\subsection{Datasets and Implementation Details}
We train on MNIST, Fashion-MNIST~\cite{xiao2017fashion}, CIFAR-10~\cite{krizhevsky2009cifar}, and SVHN~\cite{netzer2011svhn} with standard splits. For CapsNet and IBCapsNet, we use the original architecture with 32 primary capsules and 16-dimensional digit capsules. The IB regularization weight $\beta$ is tuned per dataset ($\beta=10^{-3}$ for MNIST/Fashion-MNIST, $10^{-2}$ for SVHN/CIFAR-10). Noise levels are uniformly sampled from $[0, 0.9]$ for clamped-additive/multiplicative/salt-pepper noises, and blur kernel $\sigma \in [0, 3]$ for gaussian blur.

\subsection{Clean Accuracy Comparison}
Table~\ref{tab:clean_acc} reports peak test accuracy under clean conditions. IBCapsNet achieves performance on par with CapsNet across all datasets, with at most a 0.05\% drop on MNIST and Fashion-MNIST, and a 0.11\% drop on SVHN—demonstrating that IB regularization does not compromise representation fidelity. IBCapsNet achieves substantial speedups in both training and inference, while reducing model parameters, demonstrating superior computational efficiency over CapsNet.
\begin{table}[!t]
\centering
\caption{Test accuracy (\%) on clean data}
\label{tab:clean_acc}
\begin{tabular}{cccc}
\hline
Dataset & LeNet & CapsNet & IBCapsNet \\
\hline
MNIST & 98.99 & 99.46 & 99.41 \\
F-MNIST & 90.17 & 90.83 & 90.78 \\
SVHN &  85.75 & 92.12 & 92.01 \\
\hline
\end{tabular}
\end{table}

\begin{table}[!t]
\caption{Performance Comparison of CapsNet vs. IBCapsNet. All experiments are conducted on the MNIST dataset with identical hardware (A100) and batch size (128).}
\label{tab:capsnet_vs_ibcapsnet}
\begin{tabular}{cccc}
\hline
\textbf{Metric} & \textbf{CapsNet} & \textbf{IBCapsNet} & \textbf{Improvement/Reduction} \\
\hline
Training(s/epoch) & 49.95 & 19.67 & \textbf{2.54$\times$ Faster} \\
Inference(FPS) & 41.15 & 149.93 & \textbf{3.64$\times$ Higher} \\
Parameters & 8,215,568 & 7,832,929 & \textbf{4.66\% Reduction} \\
\hline
\end{tabular}
\end{table}

The computational efficiency gains in Table~\ref{tab:capsnet_vs_ibcapsnet} stem from eliminating the iterative routing procedure. CapsNet requires 3-5 iterations per forward pass, while IBCapsNet performs a single forward pass through the global context encoder and parallel VAEs, with no iterative updates.

\begin{figure}[pt]
    \centering
    \includegraphics[width=\linewidth]{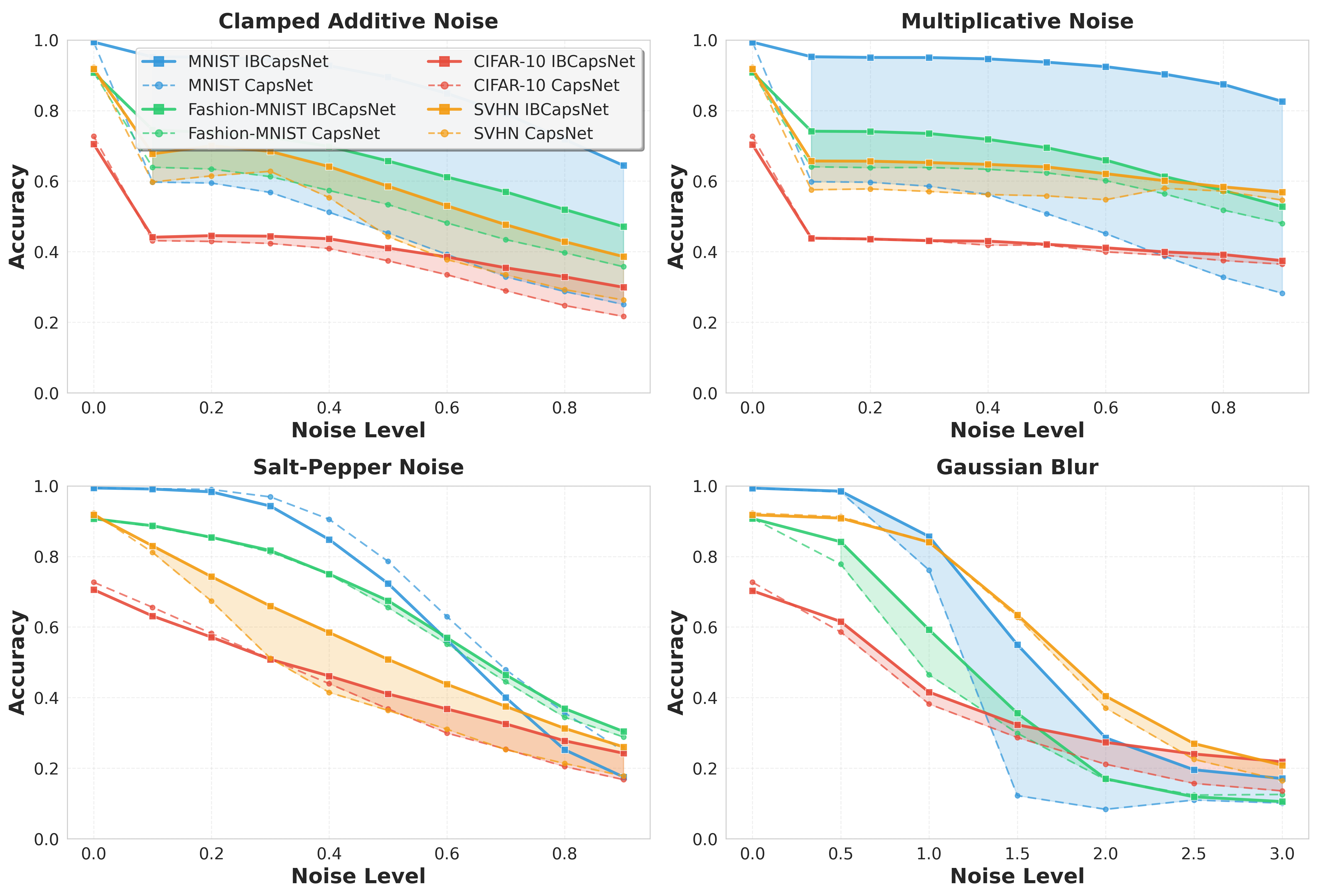}
    \caption{Robustness comparison between IBCapsNet and vanilla CapsNet across multiple datasets and disturbance scenarios.}
    \label{fig:robustness}
\end{figure}

\subsection{Robustness Under Corruptions}
We evaluate robustness under four corruptions: Clamped Additive Noise, Multiplicative Noise, Gaussian Blur, and Salt-Pepper Noise. As summarized in Table~\ref{tab:robustness_summary}, IBCapsNet consistently outperforms CapsNet, with the largest gains in \textit{Clamped Additive Noise} (+17.10\% avg.) and \textit{Multiplicative Noise} (+14.54\% avg.). These noise types are particularly challenging for CapsNet because they directly corrupt input intensity values, disrupting primary capsule activations and breaking the consensus required for dynamic routing. The information bottleneck in IBCapsNet naturally filters out these intensity-based corruptions by compressing the representation, forcing the model to rely on structural patterns rather than absolute intensity values. Notably, on MNIST, IBCapsNet improves by over 40\% in both scenarios.

\begin{table}[!t]
\centering
\caption{Average accuracy improvement (\%) of IBCapsNet over CapsNet under noise}
\label{tab:robustness_summary}
\begin{tabular}{cccccc}
\hline
Noise Type & Clamped & Multiplicative & Gaussian Blur & Salt-Pepper \\
\hline
Avg. Gain & +17.10 & +14.54 & +6.50 & +2.57 \\
\hline
\end{tabular}
\end{table}

Per-dataset analysis (see Fig.~\ref{fig:robustness}) reveals consistent trends: On \textbf{MNIST}, IBCapsNet gains +40.99\% (Clamped) and +44.06\% (Multiplicative). On \textbf{Fashion-MNIST}, gains exceed +11\% under Clamped Noise. On \textbf{SVHN}, IBCapsNet shows strong gains across \textit{all} noise types (+9.87\% to +11.17\%). Even on \textbf{CIFAR-10}, IBCapsNet maintains advantages under Blur (+5.45\%) and Clamped Noise (+4.33\%). The varying improvement magnitudes can be understood through information theory: intensity-based corruptions (clamped additive, multiplicative) are effectively filtered by the bottleneck, while spatial corruptions (blur, salt-pepper) require more abstract representations, resulting in smaller but still significant gains.

\begin{figure}[pt]
    \centering
    \includegraphics[width=\linewidth]{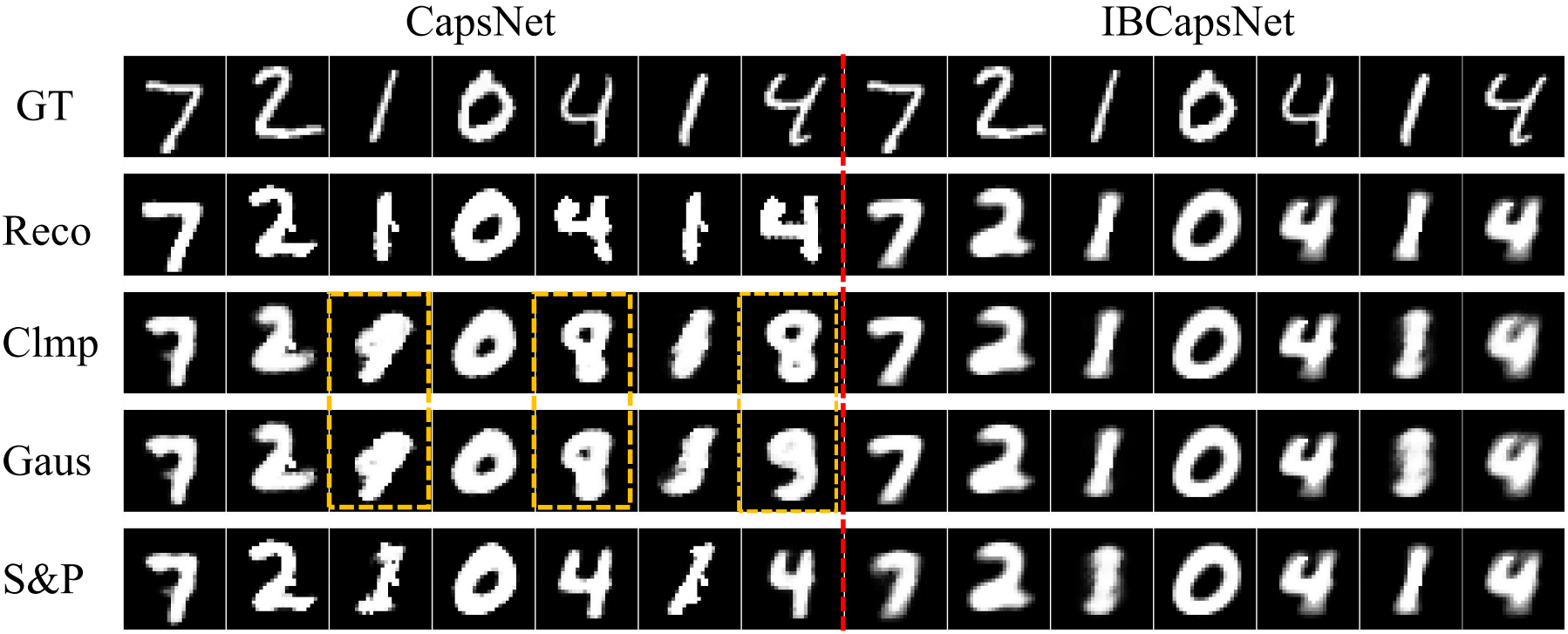}
    \caption{This figure presents the qualitative recognition performance of CapsNet (left columns) and IBCapsNet (right columns) on handwritten digits, under various disturbances. GT: Ground Truth; Reco: Recognition results without any disturbance; Clmp: adding Clamped Additive Noise; Gaus: applying Gaussian Blur; S\&P: adding Salt-Pepper Noise.}
    \label{fig:mnist_recon}
\end{figure}

\subsection{Reconstruction Visualization}
To qualitatively assess representation stability, we visualize input reconstructions from capsule outputs under increasing noise. As shown in Fig.~\ref{fig:mnist_recon}, IBCapsNet produces reconstructions that remain structurally consistent even at high noise levels, preserving semantic content with clear boundaries and recognizable shapes, with edges that remain smooth throughout. In contrast, CapsNet reconstructions degrade rapidly, exhibiting semantic shifts (e.g., digit ``4'' reconstructed as ``8'') and texture artifacts. Notably, in the second and third columns, digits ``2'' and ``1'' reconstructed by CapsNet show spurious noise artifacts (e.g., spurious spikes), while IBCapsNet maintains smooth edges. This confirms that IBCapsNet learns representations that are not only accurate but also \textit{semantically stable} under perturbation. The stability stems from the information bottleneck: by forcing the latent capsules to compress information, the model learns to encode only essential structural features needed for both classification and reconstruction, discarding noise-corrupted details.

\begin{table}[t]
\centering
\caption{Ablation study on Fashion-MNIST. Impact of reconstruction and information bottleneck on model robustness under the clamped additive noise ($\sigma = 0.3$).}
\label{tab:ablation}
\begin{tabular}{ccc}
\hline
Model Variant & Acc. (\%) & $\Delta$ vs. Baseline \\
\hline
Baseline             & 58.43 & — \\
Multi-Classifier     & 59.62 & +1.19 \\
+ Squash \& KL          & 61.55 & +3.12 \\
\textbf{+ reconstruction} & \textbf{71.71} & 
\textbf{+13.28} \\
\hline
\end{tabular}
\end{table}

\subsection{Ablation Study}
We conduct ablation studies on Fashion-MNIST under clamp noise to quantify the contribution of each component. As shown in Table~\ref{tab:ablation}, all variants achieve comparable accuracy on clean data ($\sim$90\%), confirming that architectural modifications do not compromise baseline performance. Crucially, adding the reconstruction module yields the largest robustness gain: it improves accuracy under clamp noise ($\sigma=0.3$) by \textbf{+10.16\%}, far exceeding the gains from multi-classification (+1.19\%) or KL-regularized latent capsules alone (+3.12\%). This dramatic improvement stems from the synergy between reconstruction and the information bottleneck: under the KL constraint, reconstruction acts as a denoising signal that forces latent capsules to retain only class-discriminative features sufficient for both classification and input recovery, while discarding noise-corrupted details. In contrast, CapsNet also uses reconstruction, but without an explicit bottleneck its routing-based capsules preserve high-dimensional, noise-sensitive representations, causing reconstruction to serve merely as a weak auxiliary task rather than a robustness mechanism.

\section{Conclusion}
\label{sec:conclusion}
We proposed IBCapsNet, a noise-robust capsule network grounded in the Information Bottleneck principle that addresses the poor robustness of dynamic routing to input corruptions. By replacing iterative routing with one-pass variational aggregation through class-specific VAEs regularized by KL divergence, IBCapsNet learns compact representations that naturally filter noise. Experiments on MNIST, Fashion-MNIST, SVHN, and CIFAR-10 demonstrate strong robustness under four noise types, with average improvements of +17.10\% and +14.54\% for clamped additive and multiplicative noise, while maintaining comparable accuracy on clean data. This work focuses on improving noise robustness rather than achieving state-of-the-art accuracy, and our primary contribution is a principled approach to noise-robust representation learning. The information bottleneck provides a principled mechanism for noise filtering that is fundamentally different from agreement-based routing, enabling both computational efficiency and inherent robustness to input corruptions.

\end{document}